\documentclass[10pt, a4paper]{article}
\usepackage{lrec-coling2024} 
\usepackage{amssymb}
\usepackage{natbib}
\usepackage{multibib}
\usepackage{booktabs,multirow}
\makeatletter
\def\@mb@citenamelist{cite,citep,citet,citealp,citealt,citepalias,citetalias}
\makeatother
\newcites{languageresource}{~}

\usepackage{graphicx}
\usepackage{tabularx}
\usepackage{soul}
\usepackage{threeparttable}
\newcommand{\hyperfootnote}[1][]{\def\ArgI\hyperfootnoteRelay}

\newcommand\hyperfootnoteRelay[2][]{\href{#1#2}{\ArgI}\footnote{\href{#1#2}{#2}}}

\usepackage{xcolor}
\usepackage{hyperref}
\definecolor{darkblue}{rgb}{0, 0, 0.5}
\hypersetup{colorlinks=true, citecolor=darkblue, linkcolor=darkblue, urlcolor=darkblue}

\usepackage{xstring}
\usepackage{amsmath}
\usepackage{color}

\newcommand{\concat}{\oplus}

\title{Bridging Textual and Tabular Worlds for Fact Verification: \\ A Lightweight, Attention-Based Model}

\name{Shirin Dabbaghi\textsuperscript{1},
    Canasai Kruengkrai\textsuperscript{2},
    Ramin Yahyapour\textsuperscript{1},
    Junichi Yamagishi\textsuperscript{2}}

\address{\textsuperscript{1}Gesellschaft für wissenschaftliche Datenverarbeitung mbH Göttingen (GWDG), Germany \\
    \textsuperscript{2}National Institute of Informatics (NII), Japan\\
    sdabbag@gwdg.de, canasai@gmail.com, ramin.yahyapour@gwdg.de,  jyamagis@nii.ac.jp\\}

\abstract{FEVEROUS is a benchmark and research initiative focused on fact extraction and verification tasks involving unstructured text and structured tabular data. In FEVEROUS, existing works often rely on extensive preprocessing and utilize rule-based transformations of data, leading to potential context loss or misleading encodings. This paper introduces a simple yet powerful model that nullifies the need for modality conversion, thereby preserving the original evidence's context. By leveraging pre-trained models on diverse text and tabular datasets and by incorporating a lightweight attention-based mechanism, our approach efficiently exploits latent connections between different data types, thereby yielding comprehensive and reliable verdict predictions. The model's modular structure adeptly manages multi-modal information, ensuring the integrity and authenticity of the original evidence are uncompromised. Comparative analyses reveal that our approach exhibits competitive performance, aligning itself closely with top-tier models on the FEVEROUS benchmark.
    \\ \newline  }

\begin{document}
    \maketitleabstract
    \section{Introduction}
    The proliferation of the Internet and social media has led to an increase in potentially and deliberately misleading and inaccurate statements. With minimal constraints on sharing information, anyone can now effortlessly spread erroneous or biased statements to a wide and diverse audience \cite{Saeed2021a}. Fact-checking has emerged as a crucial task to assess the veracity of claims and assertions, ensuring the dissemination of accurate and reliable information \cite{Bouziane2021, Trokhymovych, Sathe2021}.
    \par
    Several benchmark datasets have been developed to evaluate the performance of fact-checking systems. While some focus solely on text, such as Fact Extraction and VERification (FEVER) \cite{Thorne2018}, others center on tabular data, like TabFact and INFOTABS \cite{Able2019, Gupta2020}. Nevertheless, there is a pressing need for more comprehensive and inclusive benchmarks that reflect real-world fact-checking scenarios. Addressing this concern, \citet{Aly2021} introduced FEVER Over Unstructured and Structured information (FEVEROUS). FEVEROUS includes textual and tabular data to provide a more holistic representation of fact-checking tasks. In this benchmark, models are challenged to extract relevant evidence sentences or table cells from millions of unstructured passages and to incorporate this multi-modal information effectively to verify a given claim.
    \par
    Previous works on FEVEROUS have often addressed the fact verification challenge by converting all pieces of evidence into a unified format, either plain text or several tables. However, such format conversions have downsides, leading to a loss of rich context information from the original evidence and often misleading information encoding \cite{Hu2022}.
    \par
    Inspired by the work of \citet{Kruengkrai2021}, we herein propose a straightforward model that aims to eliminate the need for extensive preprocessing or rule-based requirements to convert between textual and tabular data formats. Designed with a simple modular structure, our model leverages the strengths of existing models pre-trained on tabular and text datasets and even fine-tuned for different tasks to obtain the contextual embedding of each data type. Additionally, we utilize a lightweight attention-based mechanism to explore and capitalize on the latent connections between data types. This facilitates a more comprehensive and reliable prediction of claim veracity, utilizing the strengths of both data modalities without compromising the integrity of the original evidence. A comparative analysis against existing methods reveals that our model performs competitively, achieving results nearly equivalent to the top-tier models evaluated on the FEVEROUS benchmark.

        \begin{figure*}[!ht]
            \centerline{\includegraphics[width= 0.65\textwidth]{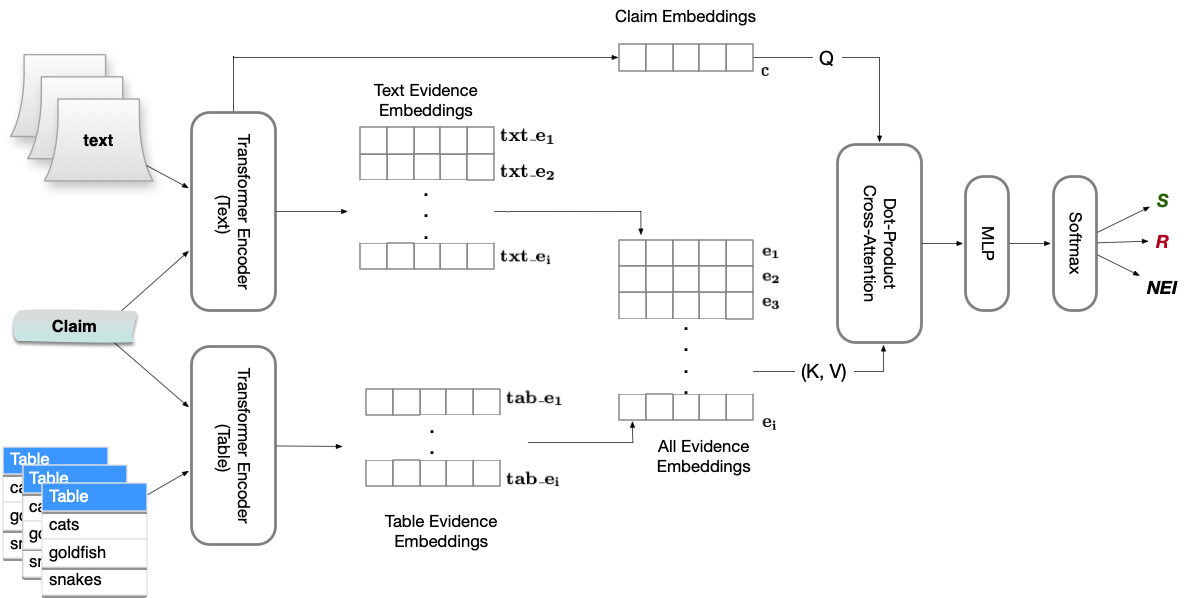}}
            \vspace{-4mm}
            \caption{The proposed model.} 
        \label{fig:model}
        \vspace{-2mm}
    \end{figure*}


    \vspace{-3mm}
    \section{Related Work}\label{RB}
    \vspace{-3mm}
    \subsection{FEVER and FEVEROUS}
    The task of fact-checking has gained prominence as an essential process for validating the veracity of claims and for promoting the circulation of reliable information \cite{Bouziane2021, Trokhymovych, Sathe2021}. A series of benchmark datasets and tasks have been developed to facilitate the advancement of automatic fact-checking algorithms.
    While each of these benchmarks has substantially contributed to the development of fact-checking models, they are often limited by the type of data they present: text-only, as in FEVER \cite{Thorne2018}, or table-only, as in TabFact, INFOTABS, and SEM-TABFACTS \cite{Able2019, Gupta2020, Wang2021}. These singular foci fail to encapsulate the diverse and multi-modal nature of real-world fact-checking scenarios. This has led to the introduction of FEVEROUS, a benchmark that combines both textual and tabular data, presenting a more nuanced challenge for fact-checking algorithms \cite{Aly2021}.
    \par
    Fact verification tasks often include two steps: evidence extraction and verdict prediction. In this research, we focus on the latter.
    \par
    In the FEVEROUS benchmark, previous works have generally relied on transforming the multi-modal evidence—text and tables—into a unified text format \cite{Aly2021, malon-2021-team} or into a unified table format \cite{Bouziane2021}.
    While this simplifies the problem to some extent, the process often results in the loss of context or nuance that is critical for accurate verdict prediction. Such an approach might limit the model's performance due to misleading information encoding.
    \citet{Hu2022} introduced a dual-channel unified format fact verification model (DCUF) in which data conversion happens in both directions simultaneously. Despite achieving a notable performance in the FEVEROUS task, their solution introduces new challenges, such as intensified preprocessing requirements, potential data redundancies, and increased computational demand.

    \vspace{-3mm}
    \subsection{Tabular Encoders}
    This subsection provides an overview of the three advanced neural tabular encoders --- TAPAS, Tapex, and Pasta --- we used in this research for proper understanding and handling of structured tabular data.

    TAPAS \cite{Herzig2020}  is a weakly supervised model for parsing and extracting information from tabular data. Unlike typical question-answering systems, TAPAS does not require manually annotated logical forms for training. It extends bidirectional encoder representations from transformers (BERT) with additional structure-aware positional embeddings to represent tables, and it is suitable for tasks like table-based question answering and fact verification, where structured data are prevalent.
    Tapex \cite{Liu2022} is a model proposed for table pre-training. It demonstrates that table pre-training can be achieved by learning a neural SQL executor over a synthetic corpus. Tapex focuses on training neural networks to understand and query tables effectively without needing real data. However, Pasta \cite{Gu2022} focuses on numerical and statistical data within tables. Pre-trained on diverse tabular datasets, Pasta can reason using statistical operations, aggregations, and numerical relationships in tables. It is ideal for tasks involving numerical computations, value prediction, and verifying claims based on statistical data.

    \begin{table*}[!ht]
        \small
        \centering
        \scalebox{0.88}{
        \begin{tabular}{@{}lccccc@{}}
            \toprule
            \multirow{2}{*}{
                \parbox[c]{.1\linewidth}{ Models}}
            & \multicolumn{2}{c}{Development} &&
            \multicolumn{2}{c}{Test} \\
            \cmidrule{2-3} \cmidrule{5-6}

            & {\centering FEVEROUS score} & {Label accuracy [\%]} && {FEVEROUS score} & {Label accuracy [\%]}  \\
            \midrule
            Official Baseline  \cite{Aly2021}       & 19            & 53                && 17.73           & 48.48 \\
            EURECOM   \cite{Saeed2021a}                & 19            &  53               &&20.01          &47.79 \\
            Z team                      &   -              &   -                  &&   22.51           &49.01 \\
            CARE        \cite{Kotonya2021}                    &  26             & 63               &&  23                     &53 \\
            NCU        \cite{Gi2021}                       &  29            &  60                &&  25.14              & 52.29\\
            Papelo              &   28                &   66            &&  25.92            & 57.57\\
            FaBULOUS  \cite{Bouziane2021}                & 30             & 65                && 27.01               & 56.07\\
            DCUF \cite{Hu2022}                          & 35.77     & 72.91         && 33.97                   & 63.21\\
            \hline
            Our Model  & 34.94& 71.86  && 32.81& 62.02\\

            \bottomrule
        \end{tabular}}
        \vspace{-2mm}

        \caption{\label{tab:comp} Model performance on the development set and test set. }
        \vspace{-2mm}
\end{table*}

\vspace{-3mm}
\section{Method}\label{mod}
Simplicity was one of our primary objectives in developing our model. Adopting a modular approach, we designed the model as a collection of independent components, each responsible for handling specific aspects of the process. This modular approach simplifies the model architecture and enables seamless adaptation to different fact-checking scenarios and datasets. Each module can be independently fine-tuned or replaced, making incorporating improvements or domain-specific enhancements easy without overhauling the entire system.
\par
For the rest of this section, we first formalize the problem domain we aim to address, followed by providing a detailed explanation of the model components and workflow.
\par
Given a claim $c$ and sets of textual evidence $TextEvids= \{ txt\_e_i\}_{i=1}^{M}$ (with $M$ being the number of pieces of text evidence) and tabular evidence $TabEvids = \{ tab\_e_i \}_{i=1}^{N}$ (with $N$ being the number of pieces of table evidence), the task is to determine the veracity of $c$. Each claim is classified into one of three categories: Supported (S), Refuted (R), or Not Enough Information (NEI).
\par
Figure \ref{fig:model} depicts an abstract overview of our solution. The model is based on a dual transformer architecture, including text and table transformers, to obtain the evidence embeddings in their original format for the given claim. For notation, we refer to these modules as $\mathrm{BERT\_text}$ and $\mathrm{BERT\_tab}$. The next section shows our solution's performance considering different transformer-based models trained for textual and tabular datasets.
\par
For a pair of a claim $c$ and $txt\_e_i$, we expect the outputted classification (CLS) token of BERT\_text to summarize the key information from $txt\_e_i$ in the context of $c$:
\begin{equation}
    \mathbf{txt\_e_i} = \mathrm{BERT\_text}_{\mathrm{CLS}}(c, txt\_e_i)
\end{equation}
Likewise, for tabular evidence $tab\_e_i$:
\begin{equation}
    \mathbf{tab\_e_i} = \mathrm{BERT\_tab}_{\mathrm{CLS}}(c, tab\_e_i)
\end{equation}
At the core of the model is a cross-attention module tasked with establishing the correlations between the textual and tabular modalities, thereby enabling the seamless fusion of their embeddings. This module aims to acquire an enriched representation of both types of evidence for the given claim, permitting a holistic and context-aware inference.
\par
The cross-attention module is based on the scaled dot-product attention mechanism introduced by \citet{Vaswani2017}. It is calculated as follows:
\begin{equation}
    Attention(Q, K, V) = softmax(\frac{QK^T}{\sqrt{d_k}})V,
\end{equation}
where $d_k$ represents the dimensionality of the keys and values used in the scaled dot-product attention mechanism. It is typically set to control the scale of the dot product.
Here, $Q$ represents the query, which is the claim representation $\mathbf{c}$, while $K$ and $V$ represent the keys and values constituted by the evidence set $E \in \mathbb{R}^{(M+N) \times D}$.
\begin{align}
    \mathbf{c} &= \mathrm{BERT\_text}_{\mathrm{CLS}}(c) \\
    E &= \{ \mathbf{e_i} \}_{i=1}^{M+N} =
    \{ \mathbf{txt\_e_i} \}_{i=1}^{M} \concat \{ \mathbf{tab\_e_i} \}_{i=1}^{N}
\end{align}

The output of the cross-attention module is a weighted sum of the evidence embeddings, creating a context-rich representation of the claim.
\par

The final stage of our model involves a classification task performed by a multi-layer perceptron (MLP) followed by a softmax layer. Let the output from the cross-attention module be denoted as \( \mathbf{z} = Attention(\mathbf{c}, E, E) \). The probability distribution, \( P \), over the three classes $[S, R, NEI]$ is then calculated as:
\begin{equation}
    P(y|\mathbf{c}, E) = \mathrm{softmax}(\mathbf{\mathrm{MLP}(\mathbf{z})})
\end{equation}

Given a ground truth label \( \mathbf{y^*} \) and the predicted probability distribution \( P(y|\mathbf{c}, E) \), we use the cross-entropy loss function to optimize the model:
\begin{equation}
    \mathcal{L} = - \sum_{i} \mathbf{y^*_i} \log P(y_i|\mathbf{c},E)
\end{equation}

\vspace{-3mm}
\section{Experiments}\label{res}
\vspace{-3mm}
\subsection{Model Settings and Configurations}

In this research, we focused on the fact verification task and relied on the retrieved evidence method presented in the work by \citep{Hu2022}\footnote{The code and model checkpoints are available at: \url{https://github.com/nii-yamagishilab/MLA-FEVEROUS-COLING24}\label{fnlabel}}. For each given claim, we considered a set of seven pieces of evidence, out of which five were in text format and the other two were tabular. Using the PyTorch framework, our model was trained on a machine with two NVIDIA Tesla A100 GPUs. We opted for the Adam optimizer with a learning rate of $1e-5$. The training was carried out for six epochs with an early stopping criterion based on the validation loss. Typically, training converged within two to three epochs, taking approximately two hours. Furthermore, we utilized an MLP comprising three layers, each followed by a ReLU activation function, and included a dropout layer with a rate of 0.2 for regularization.
\par
Our base model for the $\mathrm{BERT\_text}$ was a large DeBERTa model\footnote{\href{https://huggingface.co/MoritzLaurer/DeBERTa-v3-large-mnli-fever-anli-ling-wanli}{DeBERTa-v3-large-mnli-fever-anli-ling-wanli}} fine-tuned on multiple natural language inference (NLI) datasets. As for $\mathrm{BERT\_tab}$, we used the TAPAS large, fine-tuned specifically on the TabFact dataset\footnote{\href{https://huggingface.co/google/tapas-large-finetuned-tabfact}{tapas-large-finetuned-tabfact}}.
Furthermore, we configured the cross-attention layer with 16 attention heads and a hidden size of 1024.
\par
The experiments in this research were evaluated using a standard train-test split of the FEVEROUS dataset, in line with the evaluation metrics established for the claim verification task in this benchmark: FEVEROUS score and label accuracy in percentage.

\vspace{-3mm}
\subsection{Results and Analysis}
Table \ref{tab:comp} presents a comparative evaluation of our model against other fact-checking approaches on the FEVEROUS benchmark. Our solution achieved a FEVEROUS score of 34.94\% and a label accuracy of 71.86\%, comparable with the highest submitted model \cite{Hu2022} in the FEVEROUS benchmark.
\par
We further studied our solution using multiple transformer models fine-tuned on different datasets and tasks to show the effectiveness of the implementation design. Therefore, we evaluated the base $BERT\_tab$ (TAPAS), as well as the Tapex\footnote{\href{https://huggingface.co/microsoft/tapex-large-finetuned-tabfact}{tapex-large-finetuned-tabfact}} (400M parameters) fine-tuned on the Tabfact dataset and the Pasta large\footnote{\href{https://github.com/ruc-datalab/PASTA}{PASTA}} model with around 430M trainable parameters, which we fine-tuned on the FEVEROUS dataset. For the text encoders, we also tested RoBERTa\footnote{\href{https://huggingface.co/ynie/roberta-large-snli_mnli_fever_anli_R1_R2_R3-nli}{roberta-large-snli\_mnli\_fever\_anli\_R1\_R2\_R3-nli}} and BART\footnote{\href{https://huggingface.co/ynie/bart-large-snli_mnli_fever_anli_R1_R2_R3-nli}{bart-large-snli\_mnli\_fever\_anli\_R1\_R2\_R3-nli}}, each with a size of 350M and 400M trainable parameters, respectively.
\par
The results, presented in Table \ref{tab:models}, showcase the consistency of the overall performance in different models. Although some differences emerged, our solution structure enables any combination of text and table encoders to be used for fact-checking. In other words, in this experiment, we aimed to demonstrate just the model's versatility and robustness. As such, we consciously refrained from utilizing additional optimizations like hyperparameter tuning, ensuring that the observed performance metrics reflect the model's inherent capabilities.

\begin{table}[t]
    \centering
    \begin{threeparttable}
        \scalebox{0.88}{
        \begin{tabular}{lcc}
            \hline
            ~& FEVEROUS & Label \\
            Models & score & accuracy [\%]\\
            \hline
            TAPAS + DeBERTa &   34.94& 71.86   \\
            TAPAS + RoBERTa &   34.13 &  70.31   \\
            TAPAS + BART    &   33.85  &  69.75   \\
            Tapex + DeBERTa &   33.00  &  69.39  \\
            Tapex + RoBERTa &   32.23  &  67.27   \\
            Tapex + BART    &   31.28  &  65.38 \\
            Pasta + DeBERTa &   34.22  &  70.70 \\
            Pasta + RoBERTa &   31.60   &  66.98 \\
            Pasta + BART    &   34.53   & 70.22   \\
            \hline
        \end{tabular}}
        \vspace{-2mm}
        \caption{\label{tab:models} Model performance with different pre-trained/fine-tuned models on tabular and textual datasets.}
    \end{threeparttable}
    \vspace{-2mm}
\end{table}

\par

Table \ref{tab:evid} shows how different data types contribute to our model's performance. We evaluated both text and tabular evidence separately and in combination. When the model was trained on text-only evidence, it yielded a FEVEROUS score of 33.53\% and a label accuracy of 70.13\%. In contrast, after using table-only evidence, the scores were slightly diminished. However, both modalities propelled the model to its peak performance. This collaborative effect underscores the significance of our model's ability to seamlessly integrate textual and tabular evidence, leveraging the full potential of multi-modal data.

\begin{table}[t]
    \centering
    \begin{threeparttable}
        \scalebox{0.88}{
        \begin{tabularx}{0.9\columnwidth}{lcc}
            \hline
            ~& Feverous  & Label\\
            Data Type & score & accuracy [\%]\\
            \hline
            Only Text &  33.53&  70.13  \\
            Only Table &   32.00&  67.04  \\
            Text + Table &   34.94& 71.86  \\
            \hline
        \end{tabularx}}
        \vspace{-2mm}
        \caption{\label{tab:evid}  The importance of different data types to the model performance. 
        }
    \end{threeparttable}
    \vspace{-3mm}
\end{table}

\vspace{-3mm}
\section{Conclusion}\label{con}

The FEVEROUS task comprises two distinct sub-tasks: evidence retrieval and verdict prediction. Researchers engaging with the FEVEROUS benchmark have the flexibility to choose between these sub-tasks, allowing them to use and evaluate separate independent models when contributing to the main assignment.
In this paper, we introduced a modular, attention-based model for the verdict prediction sub-task.
\par
The proposed model aims to effectively integrate both textual and tabular data and eliminate the need for cumbersome modality conversions, often resulting in the loss of crucial context and nuance. By leveraging pre-trained models for each data type and by utilizing a lightweight cross-attention mechanism, we could effectively exploit the latent relationships between text and table data. Compared with other approaches within the FEVEROUS benchmark, our model showcases a competitive performance, underscoring its potential for accurate and versatile fact verification.

\vspace{1mm}
\noindent
\textbf{Acknowledgement:}
This work was conducted during the first author's internship at NII, Japan. This work is supported by JST CREST Grants (JPMJCR18A6 and JPMJCR20D3) and MEXT KAKENHI Grants (21H04906), Japan.


\section{References}
\vspace{-8mm}
\bibliographystyle{lrec-coling2024-natbib}
\bibliography{main}

\end{document}